\documentclass[runningheads]{llncs}

\usepackage{eccv}

\usepackage{eccvabbrv}
\usepackage[table]{xcolor}
\usepackage{xcolor}  
\usepackage{graphicx}
\usepackage{booktabs}
\usepackage{multirow}
\usepackage{wrapfig}

\usepackage[accsupp]{axessibility}  

\usepackage{hyperref}

\usepackage{orcidlink}

\begin{document}

\title{VTEdit-Bench: A Comprehensive Benchmark for Multi-Reference Image Editing  Models in Virtual Try-On} 

\titlerunning{VTEdit-Bench}

\author{Xiaoye Liang\inst{1,2}\orcidlink{0009-0004-5284-5504} \and 
Zhiyuan Qu\inst{1}\orcidlink{0009-0008-5924-9798} \and
Mingye Zou\inst{2,3} \orcidlink{0009-0002-4882-137X} \and
Jiaxin Liu\inst{1}\orcidlink{0009-0000-0642-7783} \and \\
Lai Jiang\inst{1,4}\orcidlink{0000-0002-4639-8136} \and 
Mai Xu\inst{1,4,5}\orcidlink{0000-0002-0277-3301} \and 
Yiheng Zhu\inst{2}\thanks{Corresponding author}\orcidlink{0000-0001-8020-9979}}

\authorrunning{X. ~Liang et al.}


\institute{
School of Electronic Information Engineering, Beihang University, Beijing, China
\and
Zhongguancun Academy, Beijing, China
\and
Faculty of Computing, Harbin Institute of Technology,
Harbin, China
\and
State Key Laboratory of Virtual Reality Technology and Systems, Beihang University, Beijing, China
\and
Department of Nephrology, First Medical Center of Chinese PLA General Hospital; State Key Laboratory of Kidney Diseases; and National Clinical Research Center for Kidney Diseases, Beijing, China
\\
\email{xiaoyeliang@buaa.edu.cn, zhuyiheng@zgci.ac.cn}\\
Project Page: \url{https://github.com/Hiuyee124/VTEdit-Bench}
}

\maketitle

\begin{abstract}
As virtual try-on (VTON) advances, a growing number of real-world scenarios have emerged, pushing beyond the ability of the existing specialized VTON models.
Meanwhile, universal multi-reference image editing models have progressed rapidly and exhibit strong generalization in visual editing, suggesting a promising route toward more flexible VTON systems.
However, despite their strong capabilities, the strengths and limitations of universal editors for VTON remain insufficiently explored due to the lack of systematic evaluation benchmarks.
To address this gap, we introduce VTEdit-Bench, a comprehensive benchmark designed to evaluate universal multi-reference image editing models across various realistic VTON scenarios. VTEdit-Bench contains 24,220 test image pairs spanning five representative VTON tasks with progressively increasing complexity, enabling systematic analysis of robustness and generalization. We further propose VTEdit-QA, a reference-aware VLM-based evaluator that assesses VTON performance from three key aspects: model consistency, cloth consistency, and overall image quality. 
Through this framework, we systematically evaluate 8 universal editing models and compare them with 7 specialized VTON models. Results show that top universal editors are competitive on conventional tasks and generalize more stably to harder scenarios, but remain challenged by complex reference configurations, especially multi-cloth conditioning.

  \keywords{Virtual Try-on \and Benchmark \and Multi-Reference Image Editing Models}
\end{abstract}

\section{Introduction}

\label{sec:intro}

\begin{table}[t]
\caption{Summary of supported tasks for specialized VTON models and universal multi-reference image editing models. }
  \label{tab:tab1}
  \centering
  \resizebox{\columnwidth}{!}{
  \begin{tabular}{@{}c|cccc|cccccccc|c@{}}
    \toprule
    \hline

     \multirow{3}*{\textbf{Task}} &\multicolumn{12}{c|}{\textbf{Specialized VTON Model} } & {\textbf{Universal}}\\
     \cline{2-5} \cline{6-13}
     &\multicolumn{4}{c|}{\textbf{2024} } & \multicolumn{8}{c|}{\textbf{2025} }&\textbf{Editing}\\
    &~SD.\cite{SDVTON}~&~Stable.\cite{stableviton}~&~IDM.\cite{choi2024improving}~&~TPD\cite{Yang_2024_CVPR}~&~FastFit\cite{chong2025fastfit}~&~MV.\cite{wang2025mv}~&~Omni.\cite{yang2025omnivton}~&~Any2any.\cite{guo2025any2anytryon}~&~Cat.\cite{chong2024catvton}~&~OOTDiff.\cite{xu2025ootdiffusion}~&~ITA.\cite{hong2025ita}~&~IMAG.\cite{shen2025imagdressing}~ &\textbf{Model}\\
    \midrule
    S2M & \textcolor{teal}{$\checkmark$} & \textcolor{teal}{$\checkmark$} & \textcolor{teal}{$\checkmark$} & \textcolor{teal}{$\checkmark$} & \textcolor{teal}{$\checkmark$} & \textcolor{teal}{$\checkmark$} & \textcolor{teal}{$\checkmark$} & \textcolor{teal}{$\checkmark$} & \textcolor{teal}{$\checkmark$} & \textcolor{teal}{$\checkmark$} & \textcolor{teal}{$\checkmark$} & \textcolor{teal}{$\checkmark$} & \textcolor{teal}{$\checkmark$}  \\
    S2MM&\textcolor{purple}{$\times$}&\textcolor{purple}{$\times$}&\textcolor{purple}{$\times$}&\textcolor{purple}{$\times$}&\textcolor{purple}{$\times$}&\textcolor{purple}{$\times$}&\textcolor{purple}{$\times$}&\textcolor{purple}{$\times$}&\textcolor{purple}{$\times$}&\textcolor{purple}{$\times$}&\textcolor{purple}{$\times$}&\textcolor{purple}{$\times$}&\textcolor{teal}{$\checkmark$} \\
    S2MV&\textcolor{purple}{$\times$}&\textcolor{purple}{$\times$}&\textcolor{purple}{$\times$}&\textcolor{purple}{$\times$}&\textcolor{purple}{$\times$}&\textcolor{purple}{$\times$}&\textcolor{purple}{$\times$}&\textcolor{purple}{$\times$}&\textcolor{purple}{$\times$}&\textcolor{purple}{$\times$}&\textcolor{purple}{$\times$}&\textcolor{purple}{$\times$}&\textcolor{teal}{$\checkmark$}\\
    M2M &\textcolor{purple}{$\times$}&\textcolor{purple}{$\times$}&\textcolor{purple}{$\times$}&\textcolor{purple}{$\times$}&\textcolor{purple}{$\times$}&\textcolor{purple}{$\times$}&\textcolor{teal}{$\checkmark$}&\textcolor{purple}{$\times$}&\textcolor{teal}{$\checkmark$}&\textcolor{purple}{$\times$}&\textcolor{purple}{$\times$}&\textcolor{purple}{$\times$}&\textcolor{teal}{$\checkmark$}\\
    MS2M&\textcolor{purple}{$\times$}&\textcolor{purple}{$\times$}&\textcolor{purple}{$\times$}&\textcolor{purple}{$\times$}&\textcolor{teal}{$\checkmark$}&\textcolor{purple}{$\times$}&\textcolor{purple}{$\times$}&\textcolor{purple}{$\times$}&\textcolor{purple}{$\times$}&\textcolor{purple}{$\times$}&\textcolor{purple}{$\times$}&\textcolor{purple}{$\times$}&\textcolor{teal}{$\checkmark$} \\
     \midrule
     \multicolumn{14}{l}{\footnotesize \textit{Task abbreviations:} S2M (shop-to-model), S2MV (shop-to-multi-view), S2MM (shop-to-multi-model), M2M (model-to-model), 
MS2M (multi-shop-to-model).} \\
    \hline
  \bottomrule
  \end{tabular}}
\end{table}
\begin{figure}[tb]
  \centering
  \includegraphics[width=1\textwidth]{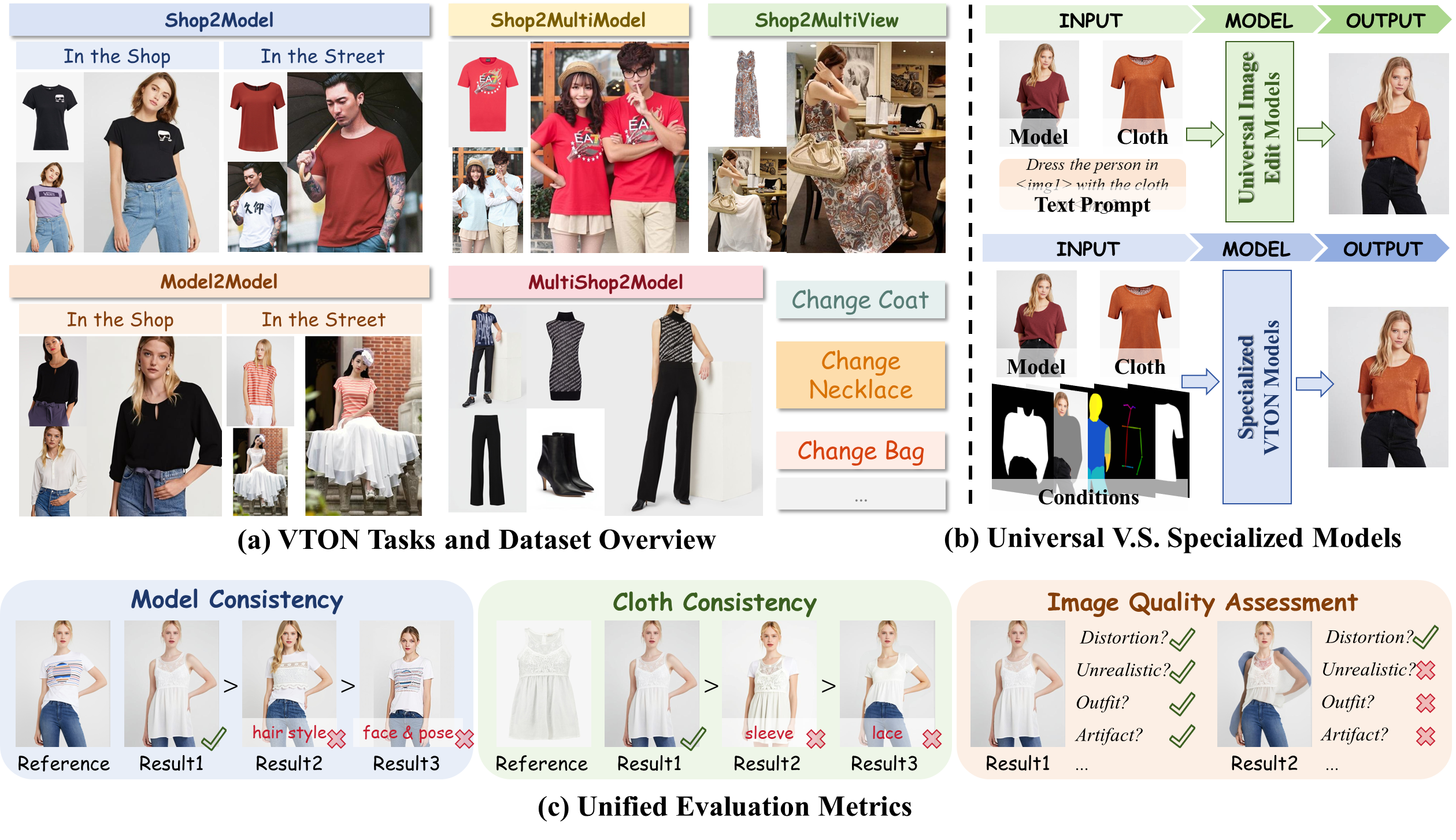}
  \caption{Overview of VTEdit-Bench, illustrating VTON tasks and dataset composition, the comparison between universal and specialized models, and the proposed unified evaluation metrics.
  }
  \label{fig:intro}
\end{figure}

In recent years, virtual try-on (VTON) has increasingly shaped online shopping and digital fashion. For consumers, it provides intuitive visualization of garments on themselves or virtual avatars, improving engagement and decision-making. For retailers, VTON enables interactive product experiences that can boost conversion and reduce returns. As e-commerce continues to expand, practical VTON systems are expected to support a broad range of real-world scenarios, as illustrated in Fig.~\ref{fig:intro}(a), from Shop2Model to Model2Model and beyond. Developing robust and adaptable VTON models is therefore crucial.

Recent progress has been driven by specialized VTON methods~\cite{SDVTON, stableviton, choi2024improving, Yang_2024_CVPR, chong2025fastfit, wang2025mv, yang2025omnivton, chong2024catvton, xu2025ootdiffusion, hong2025ita, shen2025imagdressing, guo2025any2anytryon, wan2025mft, zhang2025boow}, which largely adapt text-to-image backbones with mask-based generation or inpainting. While they achieve strong results in canonical settings such as Shop2Model, extending them to more complex VTON scenarios remains difficult. Their task-specific designs, fixed input assumptions, and reliance on auxiliary conditions often limit flexibility and scenario coverage, as summarized in Tab.~\ref{tab:tab1}.
Meanwhile, universal multi-reference image editing models~\cite{wu2025qwen, lin2025uniworld, xia2025dreamomni2, wu2025omnigen2, wu2025less, mou2025dreamo, flux-2-2025} have advanced rapidly. Following a text-conditioned image-to-image paradigm, they directly edit reference images according to instructions and naturally support multi-reference conditioning. As depicted in Fig.~\ref{fig:intro}(b), this offers a simpler and more flexible pipeline than specialized VTON methods, making universal editors well aligned with diverse VTON scenarios. This points to a potential paradigm shift, where universal multi-reference editors may evolve into a unified VTON solution and reduce reliance on task-specific pipelines. However, their effectiveness and limitations for VTON remain underexplored due to the lack of a dedicated benchmark covering realistic and diverse VTON settings. Existing VTON benchmarks \cite{li2026openvton,xiaobin2025vtbench} often focus on the most canonical and simplified try-on scenario, which is insufficient to reveal both the strengths and weaknesses of universal editors. Moreover, standard VTON evaluation still relies heavily on coarse distributional metrics (e.g., FID~\cite{FID}, KID~\cite{KID}), which overlook key factors such as model consistency, cloth consistency, and overall image quality.

To address these gaps, we introduce VTEdit-Bench, the first comprehensive benchmark for evaluating universal multi-reference editing models across diverse VTON tasks. As shown in Fig.~\ref{fig:intro}(a), 
VTEdit-Bench contains 24,220 test image pairs spanning five representative VTON settings, with task complexity progressively increasing from canonical VTON to more challenging scenarios involving viewpoint variation, identity changes, and multi-item composition.
To support fine-grained evaluation, we further propose VTEdit-QA, a reference-aware VLM-based metric that assesses VTON quality along three essential dimensions: model consistency, cloth consistency, and image quality, complementing coarse metrics and enabling more informative diagnosis. We benchmark eight universal editors against seven specialized VTON models, showing that top universal models are competitive on conventional tasks and often generalize more stably to harder settings, while challenges remain in strict identity preservation and compositional multi-item control. Overall, VTEdit-Bench provides a unified foundation for evaluating VTON systems in the era of fast-evolving universal editing models.
The main contributions are as follows:
\begin{itemize}
    \item We introduce the first comprehensive benchmark tailored for evaluating universal multi-reference image editing models under VTON scenarios with progressively increasing complexity.
    \item We introduce VTEdit-QA, a reference-aware VLM-based metric for fine-grained VTON evaluation along three key dimensions: model consistency, cloth consistency, and image quality.
    \item We benchmark eight universal editing models against seven specialized VTON models, showing comparable performance on conventional tasks and stronger generalization to novel scenarios, while leaving room for further improvement.
\end{itemize}

\section{Related Work}
\label{sec:related}

\subsection{Image-based VTON}
Image-based VTON aims to transfer a garment from a source image onto a target person image, and is widely used in online shopping. Most specialized VTON~\cite{SDVTON, stableviton, choi2024improving, Yang_2024_CVPR, chong2025fastfit, wang2025mv, yang2025omnivton, xu2025ootdiffusion, hong2025ita, shen2025imagdressing, chong2024catvton, guo2025any2anytryon} models build on text-to-image backbones, treating the person, garment, and other references as conditioning signals, and follow task-specific pipelines that are broadly mask-based or mask-free. Mask-based methods~\cite{SDVTON, stableviton, choi2024improving, Yang_2024_CVPR, chong2025fastfit, wang2025mv, yang2025omnivton, xu2025ootdiffusion, hong2025ita, shen2025imagdressing} estimate auxiliary conditions (e.g., masks/parsing/pose) and perform region-aware synthesis or inpainting, offering explicit correspondence but depending on the quality of these conditions. Mask-free methods~\cite{guo2025any2anytryon, wan2025mft, zhang2025boow} remove explicit masks and learn try-on directly from garment and person images, improving input flexibility. As VTON settings expand beyond Shop2Model to multi-view, multi-identity, and multi-item scenarios, most specialized models remain limited to a subset of tasks due to fixed input assumptions and task-specific designs.

\subsection{Instruction-based Image Editing Model}
Instruction-based image editing models~\cite{wu2025qwen, flux-2-2025} perform text-conditioned image-to-image editing, directly modifying a reference image according to a natural-language instruction. A representative early method, InstructPix2Pix~\cite{instructpix2pix}, distills supervision from large text-to-image models and trains an editor conditioned on the input image and instruction to generate the edited output.
Recent models extend this paradigm from single-image editing to multi-reference conditioning, where multiple reference images (e.g., identity, garment, or style) are jointly used with text for more controllable edits. For example, Qwen-Image-Edit-2511~\cite{wu2025qwen} supports multi-image inputs for instruction-guided editing, and FLUX.2-based editors \cite{flux-2-2025} similarly enable flexible composition of multiple visual cues within a unified generative framework. This flexibility makes universal multi-reference editors a promising alternative to specialized VTON pipelines.

\subsection{Image Editing Benchmark}
Benchmarks play a key role in assessing model capabilities. As image editing models advance, a growing number of editing benchmarks have been proposed~\cite{ye2025imgedit, jia2025compbench, pan2025ice, ma2024i2ebench, wu2025kris, pathiraja2025refedit}, including general-purpose suites that span tasks from simple to complex (e.g., I2EBench~\cite{ma2024i2ebench}, Ice-Bench~\cite{pan2025ice}, Imgedit~\cite{ye2025imgedit}) and task-specific benchmarks such as KRIS-Bench~\cite{wu2025kris} for knowledge-based reasoning edits.
VTON benchmarks have also recently emerged~\cite{cui2025street, xiaobin2025vtbench, li2026openvton}, but most focus on a single VTON setting (e.g., StreetVTON~\cite{cui2025street} for in-the-wild try-on). In contrast, we propose VTEdit-Bench to comprehensively evaluate VTON models across diverse task types, including Shop2Model, Shop2MultiModel, Shop2MultiView, Model2Model, and MultiShop2Model.

\section{VTEdit-Bench}
\label{sec:benchmark}
To evaluate universal multi-reference editors for multi-scene VTON, we propose VTEdit-Bench, which evaluates multi-scene tasks with progressively increasing reference complexity. We further introduce VTEdit-QA, a reference-aware VLM-based evaluator for fine-grained assessment of multi-reference consistency and image quality.

\subsection{Task Description}
\begin{figure}[tb]
  \centering
  \includegraphics[width=1\textwidth]{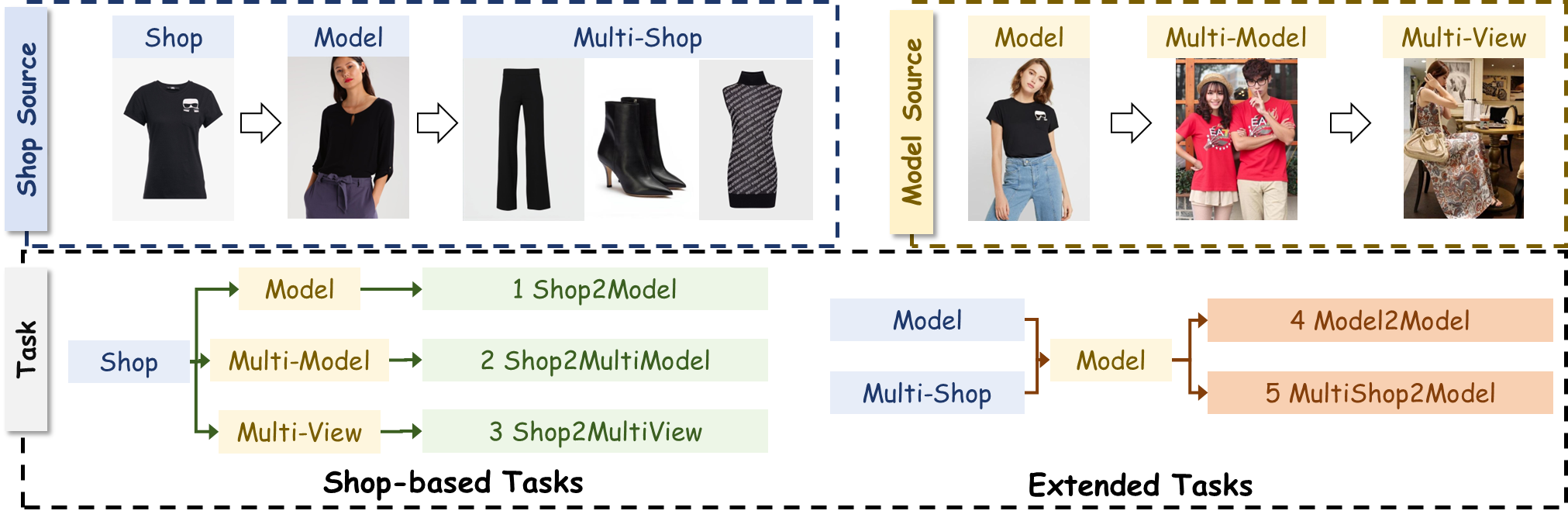}
  \caption{Illustration of the five tasks in VTEdit-Bench, showing the diversity of shop sources and model sources in the dataset, as well as how different tasks are constructed and connected through shared input compositions.}
  \label{fig:task}
\end{figure}

As shown in Fig.~\ref{fig:task}, VTEdit-Bench is organized into five representative tasks: Shop2Model, Shop2MultiModel, Shop2MultiView, Model2Model, and MultiShop-2Model. These tasks are designed to cover representative real-world VTON scenarios, including identity variation, viewpoint changes, multi-person try-on, and multi-item composition. The first three tasks are shop-based settings, where the clothing source is fixed as shop product images while the model source becomes progressively more complex, ranging from a single frontal model to multi-view and multi-person cases. The remaining two tasks extend conventional VTON beyond shop-to-person transfer by considering model-to-model garment transfer and multi-shop item composition. Overall, VTEdit-Bench comprises 13,521 human sources and 10,926 clothing sources, yielding 24,220 test pairs.

\begin{figure}[tb]
  \centering
  \includegraphics[width=1\textwidth]{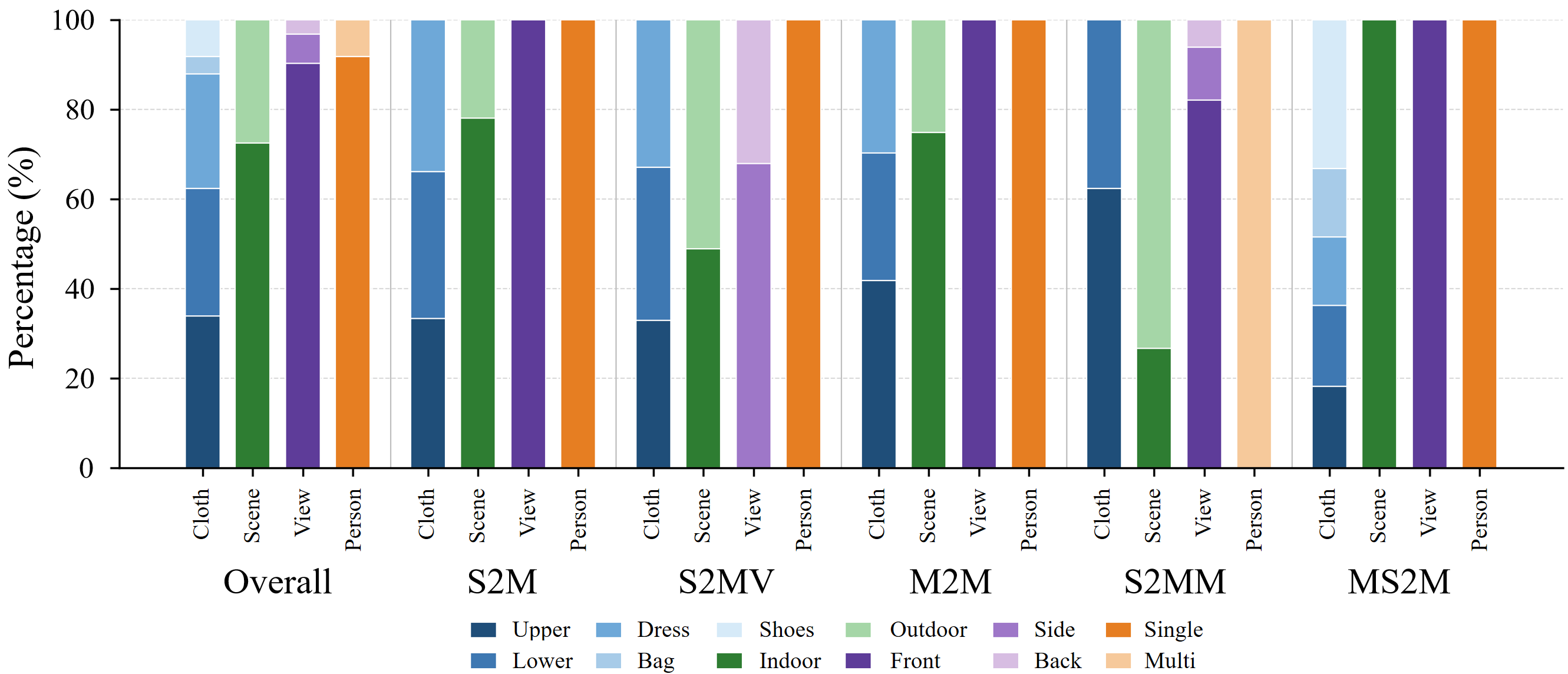}
  \caption{Attribute distribution of VTEdit-Bench across the overall benchmark and the five tasks. Each group shows the percentage distribution of four key attributes: cloth category, scene type, view direction, and person number.}
  \label{fig:dist}
\end{figure}

We further analyze the overall attribute distribution of VTEdit-Bench in Fig.~\ref{fig:dist}. The benchmark covers diverse garment categories, including upper-body clothes, lower-body clothes, dresses, bags, and shoes. It also spans different scene types, viewpoints, and person-number settings. Indoor scenes and frontal views dominate conventional settings, while outdoor scenes, non-frontal views, and multi-person cases are introduced in more challenging tasks. These distributions show that VTEdit-Bench not only covers standard VTON scenarios but also incorporates realistic variations in clothing type, scene, viewpoint, and reference composition. Details of each task are provided below.

\subsubsection{Shop2Model}
This task transfers common garments (tops, bottoms, dresses) from shop product images to model photos, evaluating realistic garment integration across indoor and outdoor scenes. Shop images and indoor models are taken from the DressCode and VITON-HD test sets, while outdoor models are collected from StreetVTON. In total, it contains 7,432 indoor models, and 2,089 outdoor models, yielding 9,521 test pairs.

\subsubsection{Shop2MultiModel}
This task extends VTON to multi-person images, evaluating consistency and coherence across individuals. We collect 400 multi-person model images from DeepFashion~\cite{deepfashion} and an additional 1,600 images from the web, and pair them with upper-body garments from the DressCode test set, following the Shop2Model setup. In total, this yields 2,000 multi-person model images and 2,000 test pairs.

\subsubsection{Shop2MultiView}
This task evaluates VTON under non-frontal viewpoints, focusing on clothing fidelity and structural coherence across perspectives. We collect 1,000 multi-view model images from DeepFashion~\cite{deepfashion} and an additional 1,000 from the web, and pair them with shop images (tops, bottoms, dresses) from the DressCode test set, following the Shop2Model setup. In total, it includes 2,000 model images and 5,400 product images, yielding 2,000 test pairs.

\subsubsection{Model2Model}
This task transfers apparel from a source person to a target person, evaluating clothing extraction and faithful re-rendering under complex appearance and background. We consider four splits: indoor $\rightarrow$ indoor, indoor $\rightarrow$ outdoor, outdoor $\rightarrow$ indoor, and outdoor $\rightarrow$ outdoor. Indoor images are from VITON-HD and outdoor images are from StreetVTON. In total, it includes 2,032 clothing sources and 4,121 model pairs, yielding 8,299 test pairs.

\subsubsection{MultiShop2Model}
This task composes multiple shop items (tops, bottoms, dresses, bags, shoes) onto a single person, stressing multi-item interaction, occlusion, and global coherence. Shop images are from the DressCodeMR test set. In total, it contains 8,894 product sets and 5,400 model images, yielding 2,400 test pairs.

\paragraph{Auxiliary conditions.}
To satisfy the input requirements of specialized VTON models, we provide auxiliary conditions for each model image, including OpenPose, human parsing, DensePose, and agnostic masks; for each clothing image, we additionally provide a cloth mask. Moreover, to ensure broad compatibility with these pipelines, we incorporate the extractability of such auxiliary signals as a constraint during data curation, filtering candidate samples to retain those for which the required conditions can be reliably computed.

\subsection{Unified Evaluation Metrics}
\label{subsec:metric}
To enable systematic and fine-grained assessment across diverse VTON tasks, we design a unified evaluation protocol. Considering the common aspects across different tasks, we design a concise, universal metric that assesses outputs along three key dimensions: model consistency, cloth consistency, and image quality.

\subsubsection{Model Consistency} This dimension evaluates whether the synthesized VTON results faithfully preserve the identity and structural attributes of the source person. Key aspects include facial identity retention, body shape consistency, and pose alignment. These criteria collectively measure the model’s ability to maintain subject-specific characteristics across diverse VTON conditions.

\subsubsection{Cloth Consistency} This dimension assesses the fidelity of garment transfer, focusing on preserving the semantic and visual attributes of the source clothing. Evaluation considers fine-grained texture details, color accuracy, and the alignment of garment boundaries with the body. Together, these factors reflect how well the model reproduces clothing characteristics across varied VTON tasks.

\subsubsection{Image Quality} This dimension evaluates the overall realism of the generated image, focusing on visual artifacts and distortions affecting the person or the clothing. Evaluation considers issues such as unnatural body proportions or deformations, distorted facial or limb details, and implausible clothing fit or fabric behavior. It also accounts for rendering artifacts (e.g., blurring or pixelation) and physical inconsistencies such as mismatched lighting or shadows.

By integrating these three dimensions, VTEdit-Bench establishes a standardized evaluation framework that jointly captures both fine-grained consistency and holistic realism. This unified design enables rigorous, reproducible, and comprehensive comparisons between specialized and universal multi-reference image editing models across a wide range of VTON tasks.

\subsection{Annotation Pipeline for Human Preferences}

\begin{figure}[tb]
  \centering
  \includegraphics[width=1\textwidth]{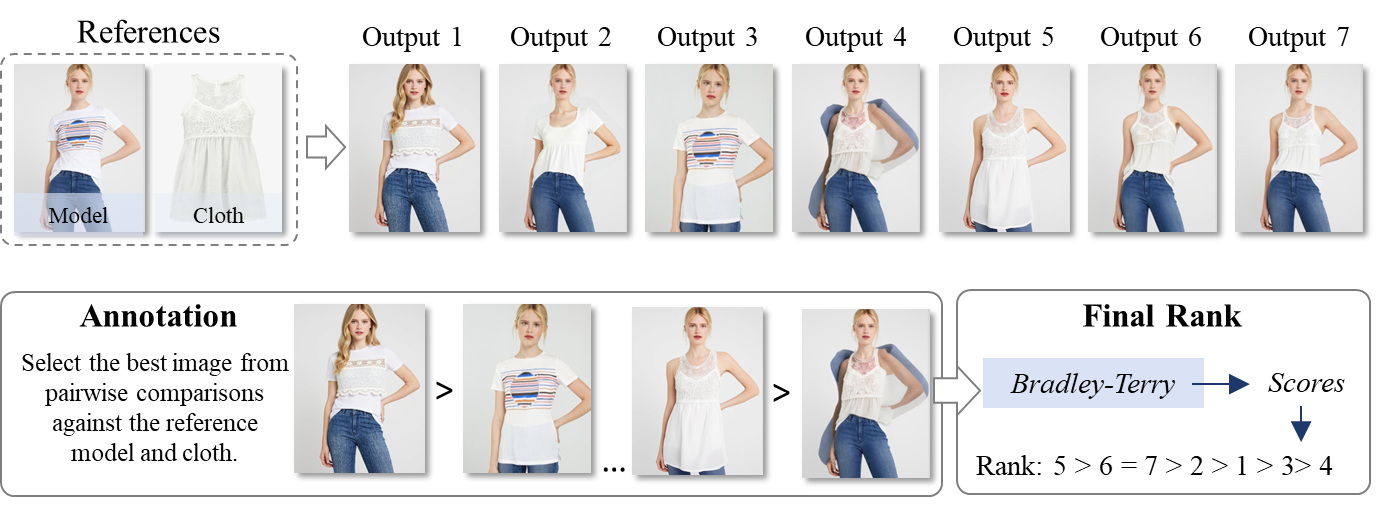}
  \caption{The annotation pipeline for capturing human preferences.}
  \label{fig:anno}
\end{figure}

To capture human preferences for VTON results, we design an annotation pipeline based on pairwise comparisons. As shown in Fig. \ref{fig:anno}, given a pair of model and cloth samples, we generate multiple outputs (output 1 to 7) using different models. The outputs are then paired, and annotators are asked to select the output with the best VTON quality. If annotators find it difficult to distinguish between the outputs, we allow them to skip the annotation. This process is repeated for all image pairs.
The pairwise comparison results are then aggregated using the Bradley-Terry model \cite{bradley1952rank} to obtain the final ranking. Let \( P_{ij} \) represent the probability that image \( i \) is preferred over image \( j \), and let \( r_i \) denote the latent "quality" or "ability" score of image \( i \). The Bradley-Terry model computes \( P_{ij} \) as:
\begin{equation}
    P_{ij} = \frac{e^{r_i}}{e^{r_i} + e^{r_j}},
\end{equation}
where \( r_i \) is the latent ranking score of image \( i \), and \( P_{ij} \) is the probability that image \( i \) is preferred over image \( j \).
The pairwise comparison results are used to estimate the values of \( r_i \) for each image using Maximum Likelihood Estimation (MLE). The final rank for each image is determined by the estimated \( r_i \) values, with higher values corresponding to higher ranks.
To ensure robust results, three annotators performed over 15k annotations, providing a comprehensive dataset of human preferences for further analysis.

\subsection{VTEdit-QA}
\begin{figure}[tb] \centering \includegraphics[width=1\textwidth]{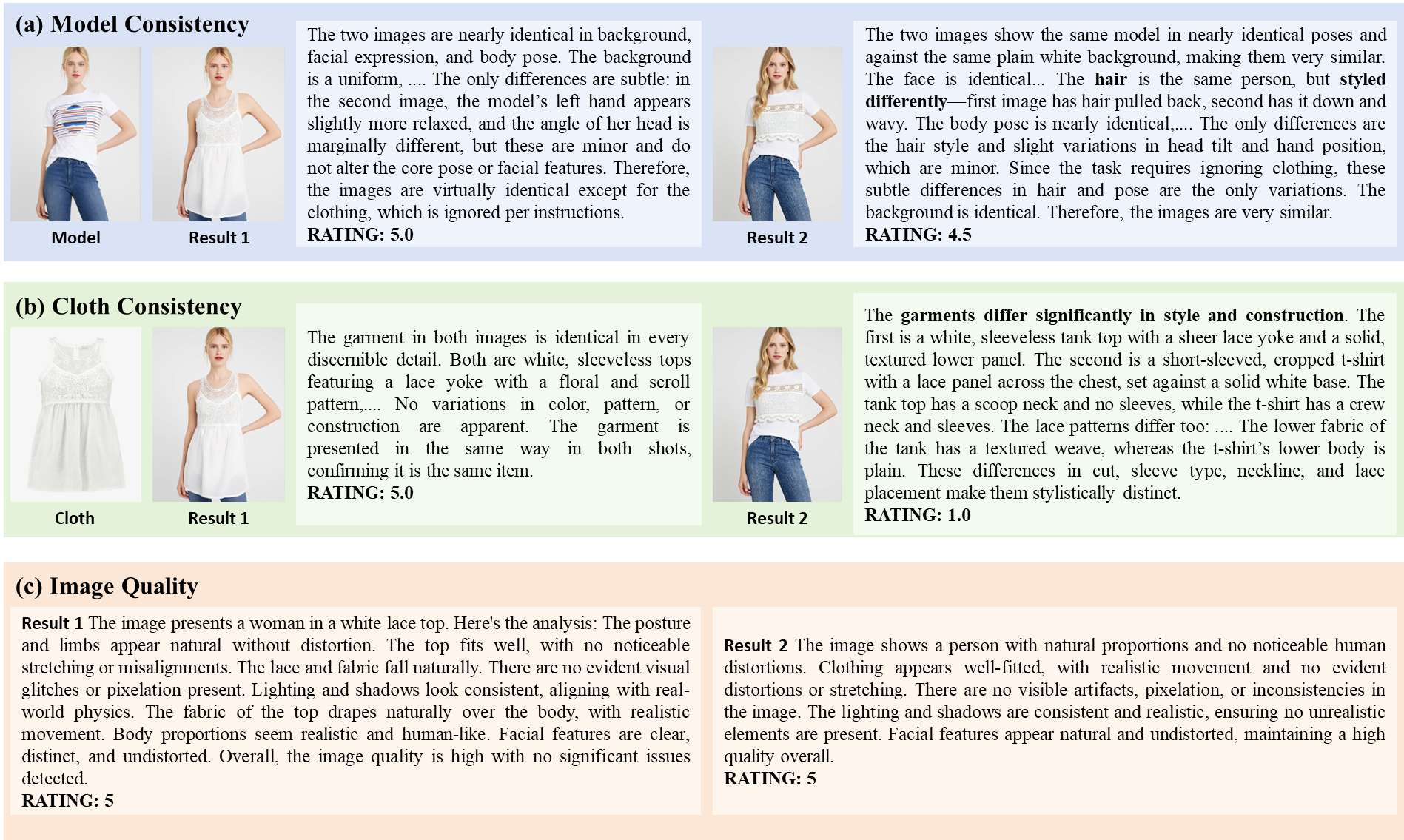} \caption{Examples of VTEdit-QA for evaluating VTON outputs along model consistency, cloth consistency, and image quality.} \label{fig:vlm} 
\end{figure}

To enable scalable evaluation, we introduce VTEdit-QA, a reference-aware quality assessment framework leveraging the large vision-language model, GPT-4o\cite{gpt4-o}, for automated scoring across three defined dimensions. Given a source model image \(S_m\), a source cloth image \(S_c\), and a generated output \(I_{out}\), GPT-4o is provided with either \(S_m\) or \(S_c\) along with \(I_{out}\) to assess model consistency and cloth consistency, respectively. For image quality, GPT-4o receives only the generated output image \(I_{out}\). For each dimension, we provide a specific prompt that details the evaluation criteria, which are further elaborated in the Supp. 1.2. Following the prompt, GPT-4o produces a numerical score between 0 and 5, reflecting the degree of consistency with the source images or the quality of the generated image. Additionally, GPT-4o generates detailed reasoning to justify its evaluation, enhancing transparency in the scoring process. 
Fig.~\ref{fig:vlm} shows two examples where GPT-4o accurately detects inconsistencies in the model and cloth, demonstrating its ability to distinguish between high-quality and low-quality outputs based on the defined evaluation criteria and providing coherent justifications for its ratings. Since a successful virtual try-on result requires simultaneously preserving the model identity, accurately transferring the garment, and maintaining overall image quality, a failure in any dimension can significantly degrade the final result. Therefore, we adopt a conservative aggregation strategy and take the minimum of the three scores as the overall score. 

\subsection{Validating VTEdit-QA via Human Correlation}
We validate VTEdit-QA by measuring how well its rankings align with human judgment. Specifically, we compare (i) the ranking consistency among human annotators and (ii) the ranking correlation between VTEdit-QA (GPT-4o) and each annotator. All correlations are computed using Spearman’s Rank Correlation Coefficient (SRCC)~\cite{spearman}, which measures the agreement between two ranked lists. Given two ranking sequences \(L_m\) and \(L_n\), SRCC is defined as:
\begin{equation}
 \text{SRCC}(L_m, L_n) = 1 - \frac{6 \sum_{j=1}^{N} d_j^2}{N(N^2-1)},
\end{equation}
where \(d_j\) denotes the rank difference of the \(j\)-th item and \(N\) is the total number of items.
\begin{figure}[tb]
\centering
\includegraphics[width=0.7\textwidth]{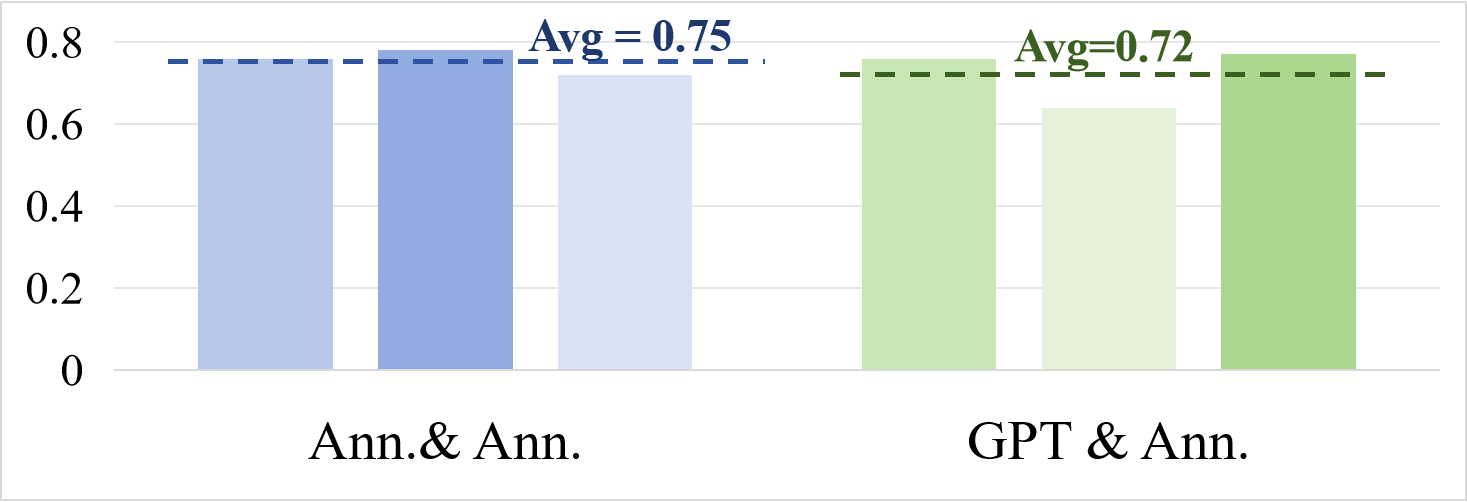}
\caption{SRCC between human annotators and GPT-4o under different scoring settings.}
\label{fig:srcc}
\end{figure}
As shown in Fig.~\ref{fig:srcc}, human annotators exhibit strong agreement, with an average Annotator--Annotator SRCC of 0.75, indicating consistent human preference rankings. Under our proposed VTEdit-QA, GPT-4o achieves an average SRCC of 0.72 with human annotator rankings, approaching the human inter-annotator agreement level. This close gap suggests that VTEdit-QA provides rankings that are largely consistent with human judgment, making GPT-4o a reliable proxy annotator for scalable evaluation. More analysis and results on variant aggregation strategies and VLMs are in Supp. 1.3.

\section{Experiment}
\begin{table}[tb]
\caption{FID and KID results on VTEdit-Bench across multiple VTON tasks. The last column reports the average of task-wise rankings for each model.}
\label{tab:kid}
\centering
\resizebox{\columnwidth}{!}{
\begin{tabular}{@{}l|cc|cc|cc|cc|cc|c@{}}
\toprule
\hline
\multirow{2}*{\textbf{Method}} &\multicolumn{2}{c|}{ ~\textbf{S2M}~ }&\multicolumn{2}{c|}{ ~\textbf{S2MV}~ } &\multicolumn{2}{c|}{ ~\textbf{S2MM}~ } &\multicolumn{2}{c|}{ ~\textbf{M2M}~ } &\multicolumn{2}{c|}{ ~\textbf{MS2M}~ } &\textbf{Avg.}\\
& FID & KID & FID & KID & FID & KID & FID & KID & FID & KID & \textbf{RANK} \\
\midrule
\rowcolor{gray!11}
\multicolumn{12}{l}{\textbf{Specialized VTON Model}}\\
IDM-VTON & 7.40 & 2.61 & 42.90 & 8.37 & 73.90 & 14.65 & - & - & - & - &\textbf{\underline{5.4}}\\
StableVITON & 16.86 & 7.24 & 37.95 & 5.01 & 111.44 & 53.58 & - & - & - & -&8.6 \\
ITA-MDT & 12.23 & 5.60 & 69.36 & 31.72 & 143.61 & 90.38 & - & - & - & -& 10.4 \\
OOTD-Diffusion & 7.49 & 2.36 & 67.94 & 22.70 & 90.57 & 26.52 & - & - & - & - & 7.6 \\
CatVTON & 10.82 & 6.95 & 43.28 & 10.54 & 76.30 & 20.03 & 14.98 & 8.79 & - & - & \textbf{\underline{5.4}}\\
Any2AnyTryon & 11.37 & 7.19 & 48.99 & 19.88 & 86.12 & 35.68 & - & - & - & - & 9.2\\
FastFit & 9.56 & 3.56 & 49.42 & 12.30 & 81.67 & 17.67 & - & - & 11.11 & 1.53 & \textbf{\underline{4.8}}\\
\hline
\rowcolor{gray!11}
\multicolumn{12}{l}{\textbf{Universal Multi-reference Image Editing Model}}\\
Qwen-Image-Edit-2511 & 9.53 & 4.17 & 43.48 & 7.88 & 71.46 & 13.14 & 13.36 & 4.43 & 46.81 & 21.60 &\textbf{\underline{3.0}} \\
Uniworld & 38.73 & 24.63 & 64.38 & 29.54 & 80.83 & 24.91 & 27.68 & 10.08 & - & - &9.2\\
Flux.2 & 10.88 & 6.54 & 36.51 & 7.26 & 65.81 & 13.68 & 14.51 & 7.62 & 15.85 & 5.36 &\textbf{\underline{2.2}}\\
DreamOmni2 & 27.41 & 14.36 & 91.79 & 45.58 & 115.67 & 43.89 & 32.36 & 12.84 & 18.77 & 6.10&9.8 \\
OmniGen2 & 34.04 & 25.42 & 81.32 & 54.24 & 83.23 & 32.47 & 37.59 & 26.68 & 35.78 & 18.61&10 \\
UNO & 32.77 & 19.87 & 70.28 & 29.97 & 93.43 & 36.24 & 29.37 & 15.63 & 53.38 & 28.71 &10.4\\
DreamO & 14.91 & 7.74 & 60.54 & 22.89 & 91.66 & 27.31 & 24.62 & 11.38 & 20.53 & 8.94& \textbf{\underline{7.4}} \\
Flux.2-klein & ~11.92~ & ~8.17~ & ~44.03~ & ~12.55~ & ~79.23~ & ~24.79~ & ~17.57~ & ~11.60~ & ~16.46~ & ~7.33~ &\textbf{\underline{5.6}}\\
 \midrule
 \multicolumn{12}{l}{\footnotesize \textit{Task abbreviations:} S2M (Shop2Model), S2MV (Shop2MultiView), S2MM (Shop2MultiModel),} \\
 \multicolumn{12}{l}{\footnotesize   M2M (Model2Model), MS2M (MultiShop2model).} \\
\hline
\bottomrule
\end{tabular}}
\end{table}

\subsection{Experimental Setup}
We benchmark universal multi-reference image editing models and compare them against specialized VTON models as strong task-specific baselines on VTEdit-Bench. Specifically, we benchmark eight open-source editing models with native multi-reference conditioning, including Qwen-Image-Edit-2511\cite{wu2025qwen}, Uniworld\cite{lin2025uniworld}, Flux.2\cite{flux-2-2025}, DreamOmni2\cite{xia2025dreamomni2}, OmniGen2\cite{wu2025omnigen2}, UNO\cite{wu2025less}, DreamO\cite{mou2025dreamo}, and Flux.2-klein\cite{flux-2-2025}. For comparison, we select seven recent open-source specialized VTON methods whose required auxiliary conditions (e.g., OpenPose, human parsing, DensePose) are supported by VTEdit-Bench, including ITA-MDT \cite{hong2025ita}, OOTD-Diffusion \cite{xu2025ootdiffusion}, CatVTON \cite{chong2024catvton}, Any2AnyTryon \cite{guo2025any2anytryon}, FastFit \cite{chong2025fastfit}, IDM-VTON \cite{choi2024improving}, and StableVITON \cite{stableviton}.
Universal multi-reference image editing models are evaluated zero-shot across all tasks using a single unified prompt (see Supp. 2.2), without any task-specific fine-tuning. Specialized VTON models are evaluated using their official checkpoints. For extended tasks (e.g., Shop2MultiModel and Shop2MultiView), we provide the required auxiliary conditions (e.g., OpenPose, human parsing) to enable zero-shot evaluation when applicable. However, most specialized VTON models are not compatible with the MultiShop2Model and Model2Model tasks due to their fixed input assumptions and task-specific requirements.
All experiments are conducted on a computing cluster with eight NVIDIA A100 GPUs to ensure consistent runtime and reproducibility.

\subsection{Preliminary Analysis on VTEdit-Bench}

As shown in Tab.~\ref{tab:kid}, we perform a preliminary analysis of universal multi-reference image editing models on VTEdit-Bench using FID and KID, where lower values indicate better realism and closer alignment to the target distribution. Here, the target distribution refers to the model images in the corresponding task-specific dataset, which serve as the reference distribution for each task.

\subsubsection{Universal Models Achieve Competitive Performance on Shop-Based Tasks.}
On the Shop2Model task, specialized VTON models achieve strong performance, led by IDM-VTON (7.40/2.61) and OOTD-Diffusion (7.49/2.36). Notably,  universal models such as Flux.2 and Qwen-Image-Edit-2511 remain highly competitive, indicating that universal multi-reference image editing models can already match strong specialized baselines in the standard VTON setting.
As task difficulty increases on Shop2MultiView and Shop2MultiModel, most methods degrade, while Flux.2 remains the best on Shop2MultiView (36.51/7.26) and strong on Shop2MultiModel (65.81/13.68), with Qwen-Image-Edit-2511 also performing competitively (43.48/7.88 and 71.46/13.14). In contrast, some specialized baselines degrade sharply, e.g., ITA-MDT rises from 12.23/5.60 to 143.61/ 90.38. These observations suggest that universal multi-reference editing models maintain more stable performance as scenario complexity increases.
\subsubsection{Universal Models Demonstrate Stronger Cross-Task Generalization.}
The extended tasks, Model2Model and MultiShop2Model, are supported by only a subset of methods, further exposing differences in model flexibility. On Model2Model, universal models such as Qwen-Image-Edit-2511 and Flux.2 achieve competitive results comparable to CatVTON. On MultiShop2Model, FastFit attains the best scores (11.11/1.53), whereas universal models show larger degradation, suggesting that fine-grained multi-item and multi-source control remains challenging for current universal editing frameworks.

In addition, some methods fail on specific tasks and produce outputs that deviate significantly from the target distribution, resulting in extremely high FID/KID values. Representative failure cases are provided in the Supp. 2.3.

\subsection{Fine-grained Analysis with VTEdit-QA}
While FID/KID measure distributional similarity, they can be misleading for VTON when clothing transfer fails (e.g., copying the input). We therefore conduct fine-grained evaluation with VTEdit-QA, which evaluates model consistency, cloth consistency, and image quality, and aggregates them into an overall score via the minimum rule. Since some methods do not support certain tasks, we rank the methods by their FID/KID scores on supported tasks, as shown in Tab.~\ref{tab:kid}, and analyze representative strong models here, including IDM, CatVTON, FastFit, DreamO, Qwen-Image-Edit-2511, Flux.2, and Flux.2-klein.

\subsubsection{Universal Models Show More Stable Performance Across Shop-based Tasks.}
As shown in Tab.~\ref{tab:part1}, we report VTEdit-QA results on the shop-based tasks supported by all evaluated methods. Across these tasks, specialized VTON models generally achieve high model and cloth consistency, but their overall scores are often limited by image quality as the settings become more challenging. In contrast, universal multi-reference editors maintain more stable overall performance across the three tasks, indicating stronger robustness to view changes and increased identity diversity. In particular, Flux.2-klein attains the highest overall scores across all three tasks (3.96/3.99/3.56), outperforming the best specialized baseline IDM (3.60/2.73/1.65).

\begin{table}[tb]
\caption{VTEdit-QA Results on shop-based tasks, including Shop2Model, Shop2MultiView, and Shop2MultiModel.}
\label{tab:part1}
\centering
\resizebox{\columnwidth}{!}{
\begin{tabular}{@{}l|cccc|cccc|cccc@{}}
\toprule
\hline
\multirow{2}*{\textbf{Method}} & \multicolumn{4}{c|}{~\textbf{Shop2Model}~} & \multicolumn{4}{c|}{~\textbf{Shop2MultiView}~} & \multicolumn{4}{c}{~\textbf{Shop2MultiModel}~} \\
& Mod. & Clo. & Qual. & Overall & Mod. & Clo. & Qual. & Overall & Mod. & Clo. & Qual. & Overall \\
\midrule
\rowcolor{gray!11}
\multicolumn{13}{l}{\textbf{Specialized VTON Models}}\\
IDM & ~4.68~ & ~4.55~ & ~4.02~ & ~3.60~ & ~4.66~ & ~4.54~ & ~3.03~ & ~2.73~ & ~4.51~ & ~4.73~ & ~1.72~ & ~1.65~ \\
CatVTON & 4.67 & 4.49 & 3.79 & 3.39 & 4.75 & 4.12 & 2.49 & 2.14 & 4.45 & 4.18 & 1.39 & 1.26 \\
FastFit & 4.64 & 4.47 & 3.42 & 3.08 & 4.70 & 4.31 & 1.94 & 1.74 & 4.45 & 4.63 & 1.22 & 1.18 \\
\rowcolor{gray!11}
\multicolumn{13}{l}{\textbf{Univeral Multi-reference Image Editing Model}}\\
DreamO & 2.78 & 3.22 & 4.19 & 1.62 & 1.82 & 3.18 & 3.42 & 0.96 & 1.62 & 2.75 & 3.07 & 0.74 \\
Qwen-Image-Edit-2511 & 4.23 & 4.77 & 4.34 & 3.64 & 3.94 & 4.20 & 3.30 & 2.46 & 4.37 & 3.74 & 2.90 & 2.14 \\
Flux.2 & 4.62 & 4.09 & 4.30 & 3.36 & 4.79 & 4.25 & 3.95 & 3.38 & 4.82 & 4.36 & 3.40 & 3.02 \\
Flux.2-klein & 4.61 & 4.68 & 4.48 & 3.96 & 4.61 & 4.80 & 4.40 & 3.99 & 4.81 & 4.73 & 3.79 & 3.56 \\
\midrule
 \multicolumn{13}{l}{\footnotesize \textit{Metric abbreviations:} Mod. (model consistency), Clo. (cloth consistency), Qual. (image quality)} \\
\hline
\bottomrule
\end{tabular}}
\end{table}
\begin{table}[tb]
\caption{VTEdit-QA Results on Model2Model and MultiShop2Model Tasks. }
\label{tab:part2}
\centering
\resizebox{\columnwidth}{!}{
\begin{tabular}{@{}l|cccc|cccccccc@{}}
\toprule
\hline
\multirow{2}*{\textbf{Method}} & \multicolumn{4}{c|}{~\textbf{Model2Model}~} & \multicolumn{8}{c}{~\textbf{MultiShop2Model}~} \\
& Mod. & Clo. & Qual. & Overall & Mod. & Dresses & Upper & Lower & Shoe & Bag & Qual. & Overall \\
\midrule
\rowcolor{gray!11}
\multicolumn{13}{l}{\textbf{Specialized VTON Models}}\\
CatVTON & 4.44 & 3.69 & 2.96 & 2.25 & - & - & - & - & - & - & - & - \\
FastFit & - & - & - & - & 3.57 & 4.96 & 4.99 & 4.95 & 4.82 & 4.23 & 4.14 & 2.90 \\
\rowcolor{gray!11}
\multicolumn{13}{l}{\textbf{Universal Multi-reference Image Editing Model}}\\
DreamO & ~1.53~ & ~3.68~ & ~4.15~ & ~0.90~ & ~2.77~ & ~2.57~ & ~3.38~ & ~2.17~ & ~0.92~ & ~1.00~ & ~4.32~ & ~0.01~ \\
Qwen-Image-Edit-2511  & 1.60 & 4.32 & 4.00 & 1.17 & 2.45 & 2.29 & 1.98 & 1.58 & 2.17 & 2.30 & 1.70 & 0.34 \\
Flux.2 & 2.77 & 3.80 & 4.36 & 2.06 & 3.18 & 4.21 & 4.82 & 4.54 & 4.20 & 4.56 & 4.20 & 2.25 \\
Flux.2-klein  & 1.55 & 2.17 & 4.63 & 1.03 & 3.52 & 4.97 & 4.65 & 4.05 & 4.56 & 4.49 & 4.45 & 2.68 \\
\midrule
 \multicolumn{13}{l}{\footnotesize \textit{Metric abbreviations:} Mod. (model consistency), Clo. (cloth consistency), Qual. (image quality)} \\
\hline
\bottomrule
\end{tabular}}
\end{table}
\subsubsection{Extended Tasks Expose Performance Gaps Among Universal Models.}
As shown in Tab.~\ref{tab:part2}, we analyze two extended tasks, Model2Model and MultiShop2Model, which are only supported by a subset of methods. Across these tasks, universal editors exhibit clear performance divergence, with some models approaching specialized baselines while others fail under the same setting.
On Model2Model, CatVTON achieves the best overall score (2.25). Among universal models, Flux.2 attains a comparable overall score (2.06), whereas Qwen-Image-Edit-2511 performs notably worse (1.17) due to low model-consistency (Mod. 1.60), suggesting frequent identity drift despite visually plausible outputs.
On MultiShop2Model, FastFit performs best (2.90 overall). Among universal editors, Flux.2-klein remains close (2.68 overall), while DreamO collapses to near-zero overall score (0.01), indicating failure to handle multi-item conditioning and composition.

\subsection{Qualitative Analysis of Universal Editing Models.}
\begin{figure}[t]
  \centering
  \includegraphics[width=\textwidth]{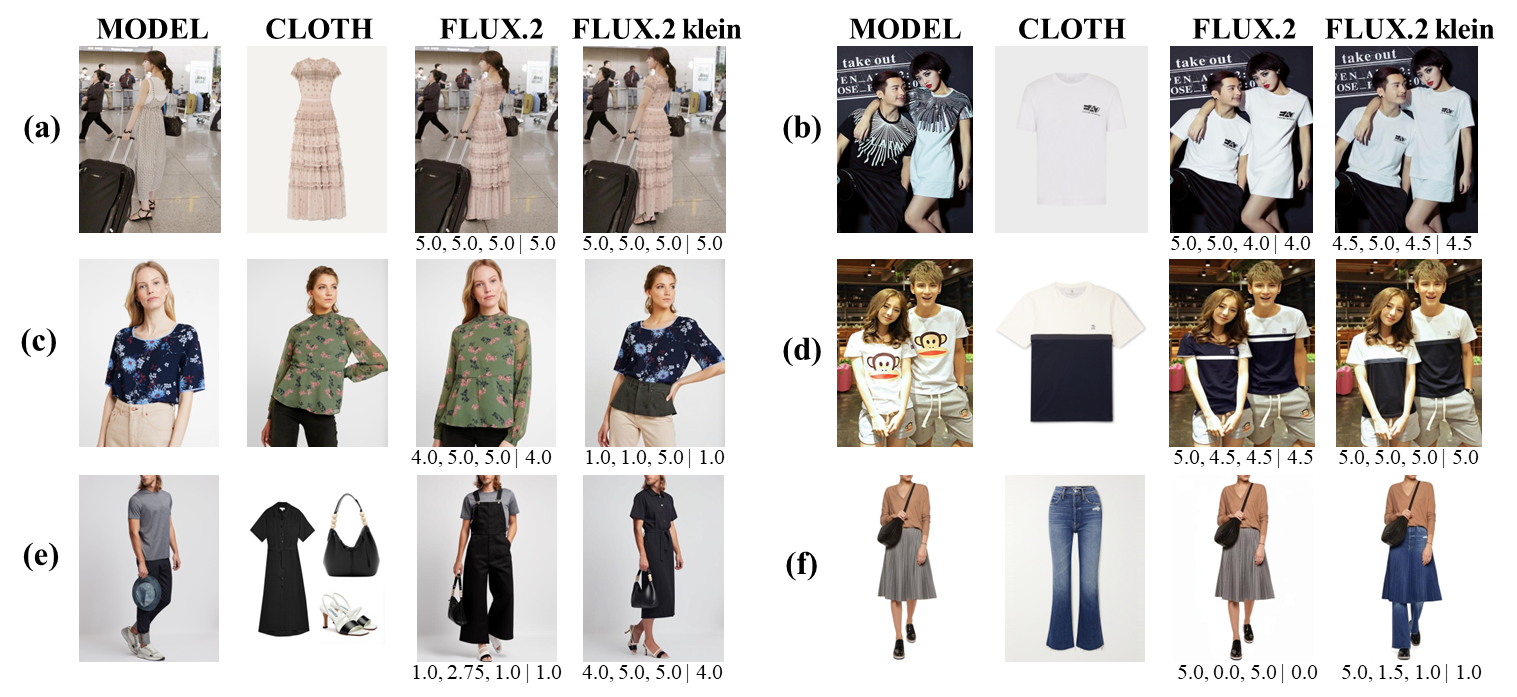}
  \caption{Qualitative examples of universal editing models (Flux.2 and Flux.2-klein) across multiple VTON tasks. Scores are reported as “model consistency, cloth consistency, image quality $\mid$ overall”.}
  \label{fig:vis}
\end{figure}

To further understand universal multi-reference editing models, we present representative qualitative cases of Flux.2 and Flux.2-klein in Fig.~\ref{fig:vis}. Fig.~\ref{fig:vis} (a,b) provide examples of successful outputs on Shop2MultiView and Shop2MultiModel, illustrating realistic synthesis and accurate garment transfer in some instances. Fig.~\ref{fig:vis} (c,d) show challenging cases where the models may confuse the person and clothing references in Model2Model, or exhibit inconsistent multi-subject control in multi-human settings, e.g., dressing only one person correctly while the other deviates from the cloth reference. Fig.~\ref{fig:vis} (e,f) further shows typical failure patterns, including unsuccessful garment replacement and incorrect garment understanding, often influenced by the source outfit. These examples qualitatively illustrate both the promise and remaining limitations of universal editors in reference understanding, identity-conditioned transfer, and robust multi-person/multi-item control.

\label{experiment}

\section{Insights and Discussions}
\label{sec:insights}

\subsection{Potential and Limitations of Universal Models} 
Based on VTEdit-Bench, universal multi-reference image editing models show clear potential for VTON. A key reason is their simpler, more general editing paradigm: instead of relying on task-specific modules, they directly perform instruction-guided editing with flexible conditioning. Together with large-scale and diverse training data, this design yields stronger generalization and more stable performance across harder variants such as Shop2MultiView and Shop2MultiModel settings, suggesting a promising path toward a unified VTON solution.
However, two bottlenecks persist: (i) model and cloth consistency degrades as reference complexity increases, and (ii) limited VTON-specific visual understanding and instruction following weaken fine-grained compositional control (e.g., attribute binding, localization, and occlusion reasoning). These findings suggest promising directions: adapting universal editors with high-quality VTON data to strengthen identity and garment constraints; improving multi-reference editing via spatially grounded reference binding and joint image–text understanding; and enabling reasoning-driven editing that iteratively analyzes failures and refines outputs toward constraint-satisfying results.

\subsection{Benchmark Scope and Future Extensions}
The current design of VTEdit-Bench uses specialized VTON models as baselines and thus primarily constructs tasks via different input image configurations to ensure fair and consistent comparison. This enables a systematic evaluation of the differences and boundaries between universal and specialized models under realistic settings.
Meanwhile, universal editing models support flexible instruction-following and prompt-driven control, opening a broader design space for VTON. Future extensions could incorporate prompt-guided settings (e.g., fine-grained constraints, multi-step instructions, localized garment replacement, and style control) to more comprehensively assess controllability and interaction capabilities.

\subsection{Evaluation Insights with VTEdit-QA}
VTEdit-QA suggests that VTON evaluation is better framed as constraint satisfaction rather than pure distribution matching. It produces rankings that are consistent with human preference and approach the level of inter-annotator agreement, indicating that the reference-aware VLM in VTEdit-QA can serve as a practical proxy for scalable assessment.
More importantly, VTEdit-QA exposes semantic failure modes that are often invisible to distributional metrics such as FID/KID. For example, IDM achieves strong FID/KID scores on VTEdit-Bench, yet receives a lower overall VTEdit-QA score than Flux.2-klein, implying that distributional similarity can remain high even when identity or garment constraints are violated. This highlights the need for VTON benchmarks to report not only realism, but also explicit reference-consistency signals (e.g., identity preservation and garment adherence) to reflect real try-on reliability.

\section{Conclusion}
\label{sec:conclusion}
We introduced VTEdit-Bench, the first benchmark for evaluating universal multi-reference image editing models under VTON scenarios with progressively increasing complexity, and benchmarked them against strong specialized baselines. Results showed that top universal editors were competitive in canonical try-on and often generalized more stably to harder multi-view and multi-model settings, while still facing bottlenecks in model/cloth consistency and compositional control.
We also introduced VTEdit-QA, a reference-aware VLM-based evaluator measuring model consistency, cloth consistency, and image quality. VTEdit-QA aligned well with human preferences and revealed semantic failures overlooked by coarse distributional metrics. Together, VTEdit-Bench and VTEdit-QA provide a unified foundation for diagnosing, comparing, and advancing VTON systems.

\section*{Acknowledgements}
This work was supported by the Zhongguancun Academy under Grants C20250402 and XTS0051, and by the National Natural Science Foundation of China under Grants 62401027 and 62231002.

\bibliographystyle{splncs04}
\bibliography{main}

@String(CVPR  = {IEEE Conf. Comput. Vis. Pattern Recog.})

@String(AAAI  = {AAAI})

@String(CVPR  = {CVPR})

@inproceedings{SDVTON,
  title={Towards squeezing-averse virtual try-on via sequential deformation},
  author={Shim, Sang-Heon and Chung, Jiwoo and Heo, Jae-Pil},
  booktitle={Proceedings of the AAAI Conference on Artificial Intelligence},
  volume={38},
  number={5},
  pages={4856--4863},
  year={2024}
}

@inproceedings{stableviton,
  title={Stableviton: Learning semantic correspondence with latent diffusion model for virtual try-on},
  author={Kim, Jeongho and Gu, Guojung and Park, Minho and Park, Sunghyun and Choo, Jaegul},
  booktitle={Proceedings of the IEEE/CVF Conference on Computer Vision and Pattern Recognition},
  pages={8176--8185},
  year={2024}
}

@inproceedings{choi2024improving,
  title={Improving diffusion models for authentic virtual try-on in the wild},
  author={Choi, Yisol and Kwak, Sangkyung and Lee, Kyungmin and Choi, Hyungwon and Shin, Jinwoo},
  booktitle={European Conference on Computer Vision},
  pages={206--235},
  year={2024},
  organization={Springer}
}

@InProceedings{Yang_2024_CVPR,
    author={Yang, Xu and Ding, Changxing and Hong, Zhibin and Huang, Junhao and Tao, Jin and Xu, Xiangmin},
    title={Texture-Preserving Diffusion Models for High-Fidelity Virtual Try-On},
    booktitle={Proceedings of the IEEE/CVF Conference on Computer Vision and Pattern Recognition (CVPR)},
    month={June},
    year={2024},
    pages={7017-7026}
}

@article{chong2025fastfit,
  title={FastFit: Accelerating Multi-Reference Virtual Try-On via Cacheable Diffusion Models},
  author={Chong, Zheng and Lei, Yanwei and Zhang, Shiyue and He, Zhuandi and Wang, Zhen and Zhang, Xujie and Dong, Xiao and Wu, Yiling and Jiang, Dongmei and Liang, Xiaodan},
  journal={arXiv preprint arXiv:2508.20586},
  year={2025}
}

@inproceedings{wang2025mv,
  title={Mv-vton: Multi-view virtual try-on with diffusion models},
  author={Wang, Haoyu and Zhang, Zhilu and Di, Donglin and Zhang, Shiliang and Zuo, Wangmeng},
  booktitle={Proceedings of the AAAI Conference on Artificial Intelligence},
  volume={39},
  number={7},
  pages={7682--7690},
  year={2025}
}

@inproceedings{yang2025omnivton,
  title={Omnivton: Training-free universal virtual try-on},
  author={Yang, Zhaotong and Li, Yuhui and He, Shengfeng and Li, Xinzhe and Xu, Yangyang and Dong, Junyu and Du, Yong},
  booktitle={Proceedings of the IEEE/CVF International Conference on Computer Vision},
  pages={16702--16711},
  year={2025}
}

@inproceedings{guo2025any2anytryon,
  title={Any2anytryon: Leveraging adaptive position embeddings for versatile virtual clothing tasks},
  author={Guo, Hailong and Zeng, Bohan and Song, Yiren and Zhang, Wentao and Liu, Jiaming and Zhang, Chuang},
  booktitle={Proceedings of the IEEE/CVF International Conference on Computer Vision},
  pages={19085--19096},
  year={2025}
}

@article{chong2024catvton,
  title={Catvton: Concatenation is all you need for virtual try-on with diffusion models},
  author={Chong, Zheng and Dong, Xiao and Li, Haoxiang and Zhang, Shiyue and Zhang, Wenqing and Zhang, Xujie and Zhao, Hanqing and Jiang, Dongmei and Liang, Xiaodan},
  journal={arXiv preprint arXiv:2407.15886},
  year={2024}
}

@inproceedings{xu2025ootdiffusion,
  title={Ootdiffusion: Outfitting fusion based latent diffusion for controllable virtual try-on},
  author={Xu, Yuhao and Gu, Tao and Chen, Weifeng and Chen, Arlene},
  booktitle={Proceedings of the AAAI Conference on Artificial Intelligence},
  volume={39},
  number={9},
  pages={8996--9004},
  year={2025}
}

@inproceedings{hong2025ita,
  title={Ita-mdt: Image-timestep-adaptive masked diffusion transformer framework for image-based virtual try-on},
  author={Hong, Ji Woo and Ton, Tri and Pham, Trung X and Koo, Gwanhyeong and Yoon, Sunjae and Yoo, Chang D},
  booktitle={Proceedings of the IEEE/CVF Conference on Computer Vision and Pattern Recognition},
  pages={28284--28294},
  year={2025}
}

@inproceedings{shen2025imagdressing,
  title={Imagdressing-v1: Customizable virtual dressing},
  author={Shen, Fei and Jiang, Xin and He, Xin and Ye, Hu and Wang, Cong and Du, Xiaoyu and Li, Zechao and Tang, Jinhui},
  booktitle={Proceedings of the AAAI Conference on Artificial Intelligence},
  volume={39},
  number={7},
  pages={6795--6804},
  year={2025}
}

@inproceedings{wan2025mft,
  title={MFT-VITON: High-Fidelity Virtual Try-On with Minimal Input via a Mask-Free Transformer-Diffusion Model},
  author={Wan, Zhenchen and Xu, Yanwu and Hu, Dongting and Cheng, Weilun and Chen, Tianxi and Wang, Zhaoqing and Liu, Feng and Liu, Tongliang and Gong, Mingming},
  booktitle={Proceedings of the IEEE/CVF International Conference on Computer Vision},
  pages={1985--1994},
  year={2025}
}

@inproceedings{zhang2025boow,
  title={Boow-vton: Boosting in-the-wild virtual try-on via mask-free pseudo data training},
  author={Zhang, Xuanpu and Song, Dan and Zhan, Pengxin and Chang, Tianyu and Zeng, Jianhao and Chen, Qingguo and Luo, Weihua and Liu, An-An},
  booktitle={Proceedings of the Computer Vision and Pattern Recognition Conference},
  pages={26399--26408},
  year={2025}
}

@article{ye2025imgedit,
  title={Imgedit: A unified image editing dataset and benchmark},
  author={Ye, Yang and He, Xianyi and Li, Zongjian and Lin, Bin and Yuan, Shenghai and Yan, Zhiyuan and Hou, Bohan and Yuan, Li},
  journal={arXiv preprint arXiv:2505.20275},
  year={2025}
}

@article{jia2025compbench,
  title={CompBench: Benchmarking Complex Instruction-guided Image Editing},
  author={Jia, Bohan and Huang, Wenxuan and Tang, Yuntian and Qiao, Junbo and Liao, Jincheng and Cao, Shaosheng and Zhao, Fei and Feng, Zhaopeng and Gu, Zhouhong and Yin, Zhenfei and others},
  journal={arXiv preprint arXiv:2505.12200},
  year={2025}
}

@inproceedings{pan2025ice,
  title={Ice-bench: A unified and comprehensive benchmark for image creating and editing},
  author={Pan, Yulin and He, Xiangteng and Mao, Chaojie and Han, Zhen and Jiang, Zeyinzi and Zhang, Jingfeng and Liu, Yu},
  booktitle={Proceedings of the IEEE/CVF International Conference on Computer Vision},
  pages={16586--16596},
  year={2025}
}

@article{ma2024i2ebench,
  title={I2ebench: A comprehensive benchmark for instruction-based image editing},
  author={Ma, Yiwei and Ji, Jiayi and Ye, Ke and Lin, Weihuang and Wang, Zhibin and Zheng, Yonghan and Zhou, Qiang and Sun, Xiaoshuai and Ji, Rongrong},
  journal={Advances in Neural Information Processing Systems},
  volume={37},
  pages={41494--41516},
  year={2024}
}

@article{wu2025kris,
  title={KRIS-Bench: Benchmarking Next-Level Intelligent Image Editing Models},
  author={Wu, Yongliang and Li, Zonghui and Hu, Xinting and Ye, Xinyu and Zeng, Xianfang and Yu, Gang and Zhu, Wenbo and Schiele, Bernt and Yang, Ming-Hsuan and Yang, Xu},
  journal={arXiv preprint arXiv:2505.16707},
  year={2025}
}

@inproceedings{pathiraja2025refedit,
  title={RefEdit: A Benchmark and Method for Improving Instruction-based Image Editing Model on Referring Expressions},
  author={Pathiraja, Bimsara and Patel, Maitreya and Singh, Shivam and Yang, Yezhou and Baral, Chitta},
  booktitle={Proceedings of the IEEE/CVF International Conference on Computer Vision},
  pages={15646--15656},
  year={2025}
}

@inproceedings{cui2025street,
  title={Street tryon: Learning in-the-wild virtual try-on from unpaired person images},
  author={Cui, Aiyu and Mahajan, Jay and Shah, Viraj and Gomathinayagam, Preeti and Liu, Chang and Lazebnik, Svetlana},
  booktitle={Proceedings of the Winter Conference on Applications of Computer Vision},
  pages={1414--1423},
  year={2025}
}

@article{xiaobin2025vtbench,
  title={VTBench: Comprehensive Benchmark Suite Towards Real-World Virtual Try-on Models},
  author={Xiaobin, Hu and Yujie, Liang and Donghao, Luo and Xu, Peng and Jiangning, Zhang and Junwei, Zhu and Chengjie, Wang and Yanwei, Fu},
  journal={arXiv preprint arXiv:2505.19571},
  year={2025}
}

@article{li2026openvton,
  title={OpenVTON-Bench: A Large-Scale High-Resolution Benchmark for Controllable Virtual Try-On Evaluation},
  author={Li, Jin and Chen, Tao and Jiang, Shuai and Wang, Weijie and Luo, Jingwen and Wu, Chenhui},
  journal={arXiv preprint arXiv:2601.22725},
  year={2026}
}

@misc{gpt4-o,
author = {OpenAI},
title = {Hello GPT-4o},
howpublished = {https://openai.com/index/hello-gpt-4o/},
note = {2024}
}

@article{wu2025qwen,
  title={Qwen-image technical report},
  author={Wu, Chenfei and Li, Jiahao and Zhou, Jingren and Lin, Junyang and Gao, Kaiyuan and Yan, Kun and Yin, Sheng-ming and Bai, Shuai and Xu, Xiao and Chen, Yilei and others},
  journal={arXiv preprint arXiv:2508.02324},
  year={2025}
}

@article{lin2025uniworld,
  title={Uniworld-v1: High-resolution semantic encoders for unified visual understanding and generation},
  author={Lin, Bin and Li, Zongjian and Cheng, Xinhua and Niu, Yuwei and Ye, Yang and He, Xianyi and Yuan, Shenghai and Yu, Wangbo and Wang, Shaodong and Ge, Yunyang and others},
  journal={arXiv preprint arXiv:2506.03147},
  year={2025}
}

@misc{flux-2-2025,
    author={Black Forest Labs},
    title={{FLUX.2: Frontier Visual Intelligence}},
    year={2025},
    howpublished={\url{https://bfl.ai/blog/flux-2}},
}

@article{xia2025dreamomni2,
  title={Dreamomni2: Multimodal instruction-based editing and generation},
  author={Xia, Bin and Peng, Bohao and Zhang, Yuechen and Huang, Junjia and Liu, Jiyang and Li, Jingyao and Tan, Haoru and Wu, Sitong and Wang, Chengyao and Wang, Yitong and others},
  journal={arXiv preprint arXiv:2510.06679},
  year={2025}
}

@article{wu2025omnigen2,
  title={Omnigen2: Exploration to advanced multimodal generation},
  author={Wu, Chenyuan and Zheng, Pengfei and Yan, Ruiran and Xiao, Shitao and Luo, Xin and Wang, Yueze and Li, Wanli and Jiang, Xiyan and Liu, Yexin and Zhou, Junjie and others},
  journal={arXiv preprint arXiv:2506.18871},
  year={2025}
}

@inproceedings{wu2025less,
  title={Less-to-more generalization: Unlocking more controllability by in-context generation},
  author={Wu, Shaojin and Huang, Mengqi and Wu, Wenxu and Cheng, Yufeng and Ding, Fei and He, Qian},
  booktitle={Proceedings of the IEEE/CVF International Conference on Computer Vision},
  pages={18682--18692},
  year={2025}
}

@inproceedings{mou2025dreamo,
  title={Dreamo: A unified framework for image customization},
  author={Mou, Chong and Wu, Yanze and Wu, Wenxu and Guo, Zinan and Zhang, Pengze and Cheng, Yufeng and Luo, Yiming and Ding, Fei and Zhang, Shiwen and Li, Xinghui and others},
  booktitle={Proceedings of the SIGGRAPH Asia 2025 Conference Papers},
  pages={1--12},
  year={2025}
}

@inproceedings{deepfashion,
  title={Deepfashion: Powering robust clothes recognition and retrieval with rich annotations},
  author={Liu, Ziwei and Luo, Ping and Qiu, Shi and Wang, Xiaogang and Tang, Xiaoou},
  booktitle={Proceedings of the IEEE conference on computer vision and pattern recognition},
  pages={1096--1104},
  year={2016}
}

@article{bradley1952rank,
  title={Rank analysis of incomplete block designs: I. the method of paired comparisons},
  author={Bradley, Ralph Allan and Terry, Milton E},
  journal={Biometrika},
  volume={39},
  number={3/4},
  pages={324--345},
  year={1952},
  publisher={JSTOR}
}

@article{spearman,
  title={The proof and measurement of association between two things.},
  author={Spearman, Charles},
  year={1961},
  publisher={Appleton-Century-Crofts}
}

@article{FID,
  title={Gans trained by a two time-scale update rule converge to a local nash equilibrium},
  author={Heusel, Martin and Ramsauer, Hubert and Unterthiner, Thomas and Nessler, Bernhard and Hochreiter, Sepp},
  journal={Advances in neural information processing systems},
  volume={30},
  year={2017}
}

@article{KID,
  title={Demystifying mmd gans},
  author={Bi{\'n}kowski, Miko{\l}aj and Sutherland, Danica J and Arbel, Michael and Gretton, Arthur},
  journal={arXiv preprint arXiv:1801.01401},
  year={2018}
}

@inproceedings{instructpix2pix,
  title={Instructpix2pix: Learning to follow image editing instructions},
  author={Brooks, Tim and Holynski, Aleksander and Efros, Alexei A},
  booktitle={Proceedings of the IEEE/CVF conference on computer vision and pattern recognition},
  pages={18392--18402},
  year={2023}
}

@article{Openpose,
  author = {Z. {Cao} and G. {Hidalgo Martinez} and T. {Simon} and S. {Wei} and Y. A. {Sheikh}},
  journal = {IEEE Transactions on Pattern Analysis and Machine Intelligence},
  title = {OpenPose: Realtime Multi-Person 2D Pose Estimation using Part Affinity Fields},
  year = {2019}
}

@misc{Densepose,
  author =       {Yuxin Wu and Alexander Kirillov and Francisco Massa and
                  Wan-Yen Lo and Ross Girshick},
  title =        {Detectron2},
  howpublished = {{https://github.com/facebookresearch/detectron2}},
  year =         {2019}
}
\end{document}